\documentclass[letterpaper]{article}
\usepackage[preprint]{aaai2027}
\usepackage[hyphens]{url}
\usepackage{graphicx}
\graphicspath{{figures/pdf/}}
\usepackage{natbib}
\usepackage{caption}
\usepackage{hyperref}
\usepackage{algorithm}
\usepackage{algorithmic}
\usepackage{amsmath,amsfonts}
\usepackage{booktabs}
\usepackage{cuted}

\usepackage{newfloat}
\usepackage{listings}
\DeclareCaptionStyle{ruled}{labelfont=normalfont,labelsep=colon,strut=off}
\floatstyle{ruled}
\newfloat{listing}{tb}{lst}{}
\floatname{listing}{Listing}

\title{AutoPref: Automatic Discovery of Task-Specific Preference Objectives for \\Neural Combinatorial Optimization}
\author{
    Shengda Gu\textsuperscript{\rm 1,2},
    Kai Li\textsuperscript{\rm 1,2},
    Xinyi Ke\textsuperscript{\rm 1,2},
    Haobo Fu\textsuperscript{\rm 3},
    Yifan Zhang\textsuperscript{\rm 4},
    Jian Cheng\textsuperscript{\rm 1,2,5}
}
\affiliations{
    \textsuperscript{\rm 1}C\textsuperscript{2}DL, Institute of Automation, Chinese Academy of Sciences\\
    \textsuperscript{\rm 2}School of Artificial Intelligence, University of Chinese Academy of Sciences\\
    \textsuperscript{\rm 3}Tencent AI Lab\\
    \textsuperscript{\rm 4}C2DL, Institute of Automation, Chinese Academy of Sciences\\
    \textsuperscript{\rm 5}AiRiA
}
\copyrighttext{Preprint. This manuscript has not been peer reviewed. The authors retain copyright and reserve all rights. This version may be updated; please cite the arXiv version when available.}
\begin{document}

\maketitle

\begin{abstract}
Combinatorial optimization problems (COPs) underpin many real-world decisions, yet their exponentially growing search spaces make high-quality solutions expensive to obtain. Neural combinatorial optimization (NCO) addresses this challenge by learning fast construction policies, typically through reinforcement learning (RL). To overcome the sample inefficiency of RL, preference-based NCO has emerged as a new paradigm, learning directly from the relative quality of sampled solutions. Fundamentally, an effective preference objective must answer two critical questions: (1) what to learn from a pair of solutions, and (2) how much to learn from each pair relative to the sampled set. Existing methods bind both answers into manually designed, one-scale-fits-all formulations, failing to capture the diverse, task-specific requirements of different COPs. In this paper, we present \textit{AutoPref}, the first LLM-guided automated discovery framework for preference-objective design for NCO, formalizing it over a coupled programmatic space. Rather than relying on human intuition, we explicitly factorize the objective into a pairwise loss program that specifies the learning signal from each pair, and a set-aware weighting program that controls its relative contribution. Their composition defines a unified objective space that includes existing preference objectives as special cases. Since jointly searching this space is prohibitively expensive, we propose an efficient staged conditional search strategy, augmented by behavioral gates that rigorously screen out behaviorally inadmissible programs prior to short-horizon downstream training and evaluation. Across TSP, CVRP, FFSP, and JSSP, the discovered objectives consistently outperform the strongest hand-designed baselines across various problem scales, highlighting the necessity and scalability of automated objective discovery over manual design for NCO.
\end{abstract}

\section{Introduction}
Combinatorial optimization problems (COPs), such as routing and scheduling, arise in a wide range of real-world applications~\citep{bengio2021machine}. Many COPs are NP-hard, with their solution spaces growing exponentially with problem scale~\citep{korte2011combinatorial}. Exact algorithms guarantee optimality but often become impractical at scale, whereas heuristic methods trade these guarantees for efficiency and typically require substantial expert design.

Neural combinatorial optimization (NCO) offers a promising alternative by learning construction policies that generate high-quality solutions efficiently~\citep{vinyals2015pointer,bello2016neural,kool2019attention}. Supervised NCO~\citep{vinyals2015pointer} requires optimal or near-optimal labels, while reinforcement learning (RL) removes this requirement and has become the dominant training paradigm~\citep{bello2016neural,kool2019attention}. However, RL-based NCO remains sample inefficient: scalar rewards become less discriminative as the policy improves, and training underuses multiple solutions sampled for the same instance~\citep{pmlr-v267-pan25e,pmlr-v267-liao25a}. Preference-based NCO has therefore emerged as a promising direction: it converts relative solution quality into direct supervision and trains the policy to favor lower-cost solutions already produced during sampling~\citep{pmlr-v267-pan25e,pmlr-v267-liao25a}.

Despite its promise, the effectiveness of preference-based NCO depends critically on its training objective. A complete preference objective must consequently make two coupled design decisions. First, it must determine how an individual comparison shapes the policy update. For example, the objective must decide how to correct the probability margin between a preferred and a less-preferred solution. Second, it must determine how training emphasis is distributed across the comparisons generated from the sampled set. Some comparisons may be more informative because the policy ranks them incorrectly or because they are distinctive relative to the remaining sampled solutions.

Existing preference-based NCO methods rely on manually designed objectives to answer these questions. For example, PO4COPs~\citep{pmlr-v267-pan25e} applies a predefined pairwise preference loss uniformly to all pairs. BOPO~\citep{pmlr-v267-liao25a} combines an objective-guided pairwise loss and selects comparisons around the best solution. Such designs have demonstrated the value of preference training for NCO, but they leave the objective design to human intuition. More importantly, an objective that is effective for one COP family may not transfer to another: routing and scheduling problems differ substantially in their solution structures and reward landscapes. Manually designing a suitable objective for each task therefore requires repeated expert effort and makes it difficult to adapt preference training to task-specific learning dynamics.

We present AutoPref, an LLM-guided framework for the automatic discovery of task-specific preference objectives for NCO. Rather than treating a preference objective as a monolithic loss formula, AutoPref factorizes it according to input scope. A \emph{pairwise loss program} receives the features of the two solutions forming one preferred–less-preferred pair and determines what the policy learns from that comparison. A \emph{set-aware weighting program} additionally observes the full sampled solution set and determines how much that comparison contributes relative to the others. Their normalized weighted composition defines a coupled programmatic objective space. This representation is expressive enough to recover existing preference objectives as special cases, while enabling AutoPref to explore beyond these hand-designed rules and discover potentially stronger, task-specific objectives.

Searching this coupled space is challenging. A direct joint search must select both programs simultaneously, and every joint candidate requires an expensive policy-training evaluation. Moreover, when a candidate performs poorly, the resulting fitness provides weak credit assignment: it is difficult to determine whether the failure arises from the pairwise loss, the weighting program, or an incompatible interaction between them. AutoPref addresses this challenge through a staged conditional search strategy. In the first stage, it searches for a pairwise loss program while assigning uniform weight to every comparison. This isolates the question of what to learn from each pair. In the second stage, the first-stage pairwise loss is fixed while AutoPref searches for a set-aware weighting program, thereby optimizing how much to learn from each pair.

To further improve search efficiency, AutoPref places behavioral screening before downstream evaluation. Candidate programs must compile successfully, return finite values and gradients, preserve the intended preference direction, remain stable under changes in objective-value scale, and produce valid non-negative comparison weights. Only behaviorally valid candidates are evaluated through short-horizon NCO training. Because the distribution of sampled comparisons evolves with the policy, each candidate is evaluated from both an early policy state and a partially trained state, reducing the risk of selecting objectives that provide only transient early-stage improvements.

We evaluate AutoPref on four representative COP families: the Traveling Salesman Problem (TSP), Capacitated Vehicle Routing Problem (CVRP), Flexible Flow Shop Problem (FFSP), and Job Shop Scheduling Problem (JSSP). For each family, we search at one problem scale, freeze the discovered objective, and train new policies with the full budget. Under matched training and inference protocols, the discovered objectives set a new state of the art for preference-based NCO by outperforming the strongest hand-designed objective in every discovery- and held-out-scale setting. Together, these results support a shift from manually specifying preference objectives to automatically discovering them as task-specific components of the NCO training pipeline. More broadly, AutoPref shows that the design of learning objectives can itself be treated as an optimization problem, enabling NCO systems to adapt not only their policies, but also the principles by which those policies are trained.

\section{Preliminaries}
\label{sec:preliminaries}

Let $\pi_\theta$ denote an autoregressive construction policy. For each COP instance $x$, the policy samples a solution set
$S_x=\{\tau^{(1)},\dots,\tau^{(N)}\}$.
Each solution $\tau^{(j)}=(a_1^{(j)},\dots,a_{T_j}^{(j)})$ is constructed sequentially, with the action at decision $t$ sampled from
$\pi_\theta(\cdot\mid a_{<t}^{(j)},x)$.
Let $\lambda_{j,t}=\log\pi_\theta(a_t^{(j)}\mid a_{<t}^{(j)},x)$
denote the log-probability of the selected action. The construction log-probability of $\tau^{(j)}$ is
$
p_j
=
\log\pi_\theta(\tau^{(j)}\mid x)
=
\sum_{t=1}^{T_j}\lambda_{j,t},
$
and
$\boldsymbol{\lambda}_j=(\lambda_{j,1},\dots,\lambda_{j,T_j})$
records its stepwise log-probabilities. Its policy entropy is accumulated over the construction process as
$
H_j
=
\sum_{t=1}^{T_j}
\mathcal H\!\left(
\pi_\theta(\cdot\mid a_{<t}^{(j)},x)
\right),
\mathcal H(\pi)=-\sum_a\pi(a)\log\pi(a).
$
Let $o_j$ denote the solution cost of $\tau^{(j)}$.

We orient every unequal-cost pair from the lower-cost trajectory to the higher-cost trajectory, obtaining the comparison pool:
\[
\mathcal{P}_x
=
\{(\tau_w^{(i)},\tau_l^{(i)})\}_{i=1}^{M_x}.
\]
Here $M_x=|\mathcal P_x|$, and
$\tau_w^{(i)}\succ\tau_l^{(i)}$
indicates that $\tau_w^{(i)}$ is preferred because it has lower solution cost. Preference-based NCO uses these comparisons to increase the likelihood of preferred trajectories relative to less-preferred ones.

\begin{figure*}[t]
\centering
\includegraphics[width=0.9\textwidth]{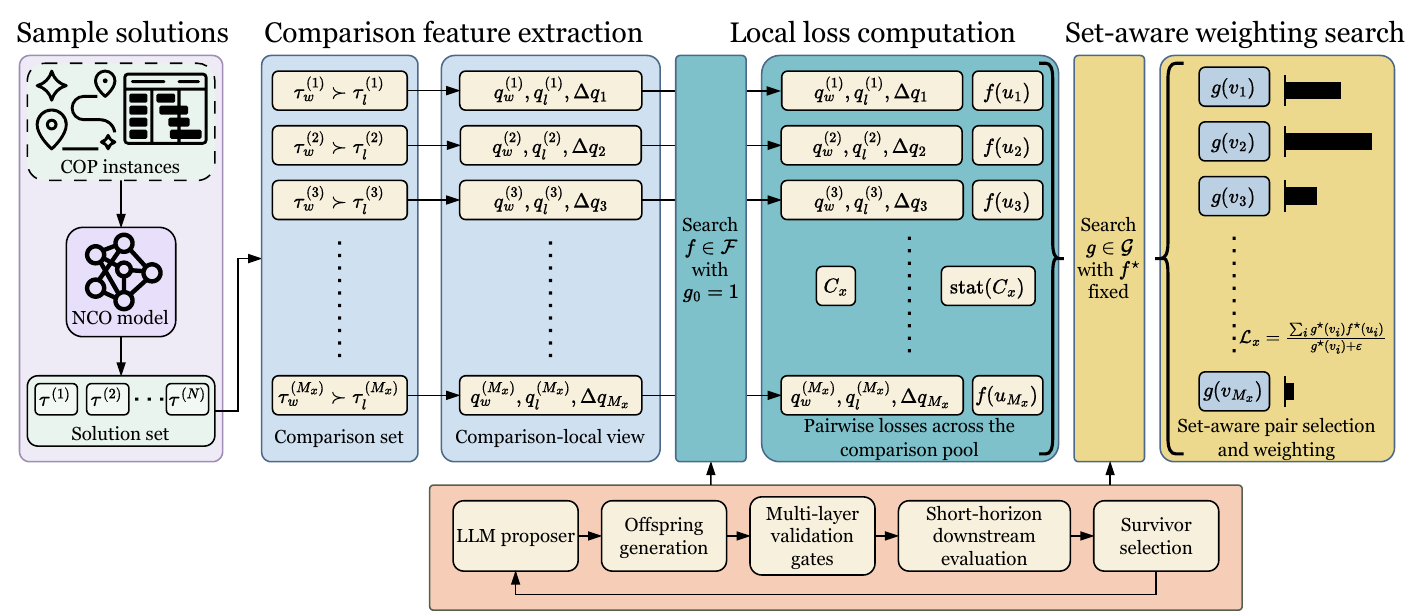}
\caption{The overall pipeline of AutoPref. An NCO policy samples a solution set, from which oriented unequal-cost comparisons and pair-level features are constructed. The first stage searches a pairwise loss program $f\in\mathcal F$ under uniform weights $g_0=1$; with the Stage-1 program $f^\star$ fixed, the second stage searches a set-aware weighting program $g\in\mathcal G$. Both stages use the same LLM-guided propose--validate--short-train--select loop, and the weighted aggregate of the resulting pairwise losses defines the discovered training objective.}
\label{fig:main-workflow}
\end{figure*}

\section{Preference Objective Space}
\label{sec:preference_objectives}
In this section, we formalize the preference objective design as a search over a programmatic space, which provides a highly expressive representation that can be efficiently explored by LLMs. Section~\ref{sec:preference_objective_discovery} will then describe how we search this space guided by downstream NCO performance.

\subsection{Preference Objective Representation}
\label{sec:preference_objective_space}

We define a preference objective as an ordered pair of executable programs $O=(f,g)$. This factorization represents two distinct algorithmic roles: the pairwise loss program $f$ dictates what to learn from a single comparison using pair-level features, while the weighting program $g$ dictates how much to emphasize that comparison using global, set-aware context.

Specifically, let $q_j=(o_j,p_j,r_j,\mathbf b_j)$ denote the feature vector attached to sampled solution $j$. Here, $r_j$ is its zero-based cost-sorted rank, with $r_j=0$ marking the best solution. The auxiliary record $\mathbf b_j$ includes the trajectory entropy $H_j$ and the stepwise log-probability sequence $\boldsymbol{\lambda}_j$ defined in Section~\ref{sec:preliminaries}. Each pairwise preference comparison provides a pair-level representation:
\[
u_i=\left(q_w^{(i)},q_l^{(i)},\Delta q_i\right),
\]
where $\Delta q_i = \left( o_l^{(i)}-o_w^{(i)}, p_w^{(i)}-p_l^{(i)}, r_l^{(i)}-r_w^{(i)} \right)$. Let $\mathcal U$ denote the domain of these pair-level representations. A pairwise loss program $f:\mathcal U\to\mathbb{R}$ returns a scalar loss term $f(u_i)$ and determines what the policy learns from comparison $i$. We denote the admissible programs by $\mathcal F$.

Conversely, the weighting program requires a broader set-contextual view:
\[
v_i=\left(u_i,\;\mathcal C_x,\;\operatorname{stat}(\mathcal C_x)\right).
\]
Here $\mathcal C_x=\{q_j\}_{j=1}^{N}$ collects the per-solution records for instance $x$. The deterministic summary $\operatorname{stat}(\mathcal C_x)$ provides precomputed instance-level quantities such as the cost median absolute deviation, and the log-probability standard deviation.
Let $\mathcal V$ denote the domain of these set-contextual views. A set-aware weighting program $g:\mathcal V\to\mathbb R_{\geq 0}$ returns a non-negative weight $g(v_i)$ that determines comparison $i$'s contribution relative to the other comparisons from $x$. 

We denote the admissible programs by $\mathcal G$. 
Appendix~A.1.4 details the specific feature design.

The programmatic preference objective space is thus defined as the Cartesian product $\mathcal F\times\mathcal G$. For any candidate objective $O=(f,g)\in\mathcal F\times\mathcal G$, the NCO policy training minimizes the expected normalized weighted loss:
\begin{equation}
\min_{\theta}\;\mathrm E_{x\sim\mathcal D}\!\left[
\frac{\sum_{i=1}^{M_x}g(v_i)f(u_i)}
{\sum_{i=1}^{M_x}g(v_i)+\varepsilon}
\right],
\label{eq:searched-objective}
\end{equation}
where \(\varepsilon>0\) is a small constant to prevent division by zero.

A key advantage of this representation is its expressiveness; it naturally unifies existing state-of-the-art preference paradigms as specific points within the $\mathcal{F}\times\mathcal{G}$ space. 
For instance, PO4COPs~\citep{pmlr-v267-pan25e} corresponds to:
\begin{equation}
f_{\mathrm{PO}}(u_i)
=-\log\phi\!\left(\alpha\Delta p_i\right),
\qquad
g_{\mathrm{PO}}(v_i)=1,
\label{eq:po4cops-special-case}
\end{equation}
where $\Delta p_i=p_w^{(i)}-p_l^{(i)}$. The preference function $\phi$ can be, for example, the Bradley--Terry sigmoid $\sigma(z)=(1+e^{-z})^{-1}$ or the exponential function $\exp(z)$. PO4COPs applies this pairwise loss uniformly to all comparisons.

Similarly, BOPO~\citep{pmlr-v267-liao25a} filters comparisons uniformly over a sampled set of scale $N$, selecting $K$ solutions with a step scale $s=\lfloor N/K\rfloor$. Its objective can be exactly reproduced as:
\begin{equation}
\begin{aligned}
f_{\mathrm B}(u_i)
&=-\log\sigma\!\left(
\frac{o_l^{(i)}}{o_w^{(i)}}
\bigl(\ell_w^{(i)}-\ell_l^{(i)}\bigr)
\right),\\
g_{\mathrm B}(v_i)
&=\mathbf 1[r_w^{(i)}=0]\,
\mathbf 1[r_l^{(i)}\bmod s=0].
\end{aligned}
\label{eq:bopo-special-case}
\end{equation}

As shown, both PO4COPs and BOPO are fundamentally hand-designed, static points within $\mathcal{F}\times\mathcal{G}$. AutoPref shifts the paradigm from manual specification to automated discovery, systematically searching this expansive space for a high-performing, task-specific program pair $(f^\star, g^\star)$.

\section{Objective Discovery}
\label{sec:preference_objective_discovery}
With the programmatic space defined, the central challenge becomes navigating it efficiently. 
To orchestrate the search, we adopt a propose-check-train-select evolutionary loop (summarized in Figure~\ref{fig:main-workflow}). In this iterative loop, an LLM acts as the generator. Previously evaluated programs and their empirical scores serve as context to guide subsequent proposals, while selected high-performing survivors form the parent set for the next generation.

We note that the high-level LLM-driven search loop is a standard technique\citep{romera2024funsearch}. The primary novelty of AutoPref lies in extending automated discovery to the design of preference objectives in NCO—a component traditionally restricted to manual specification. We intentionally utilize a simple search pipeline to demonstrate that the success of this approach is driven by our structural designs rather than complex search heuristics. Because evaluating the fitness of any candidate program inherently requires an NCO training run, naive exploration remains computationally intractable. AutoPref strategically overcomes this computational bottleneck through key mechanisms detailed in this section: staged conditional search, rigorous behavioral screening, and short-horizon downstream evaluation.

\subsection{Staged Conditional Search over the Coupled Space}

The factorization of $O=(f,g)$ makes the search target explicit but introduces a tightly coupled space $\mathcal{F}\times\mathcal{G}$. A direct joint search must select both programs simultaneously. This not only causes a combinatorial explosion in the search space but also creates a severe credit assignment problem: when a joint candidate performs poorly, it is difficult to determine whether the failure stems from an ineffective pairwise loss, a flawed weighting program, or a detrimental interaction between two.

To navigate this coupled space efficiently, we decompose the joint problem into a staged conditional search:
\begin{itemize}
    \item \textbf{Stage 1: Isolating the Learning Signal.} We first search the space $\mathcal{F}\times\{g_{0}\}$, where $g_{0}(v_{i})=1$ assigns equal weight to every comparison. The resulting Stage-1 pairwise loss program $f^\star$ defines the Uniform Pair Weighting (UPW) objective: $\mathcal{L}_{UPW}(x;f^\star)=\frac{1}{M_{x}}\sum_{i=1}^{M_{x}}f^\star(u_{i})$. By forcing uniform weights, this stage strictly isolates the optimization of what the policy learns from each individual pair.
    \item \textbf{Stage 2: Optimizing the Set-Aware Contribution.} Next, we freeze the Stage-1 loss $f^\star$ and search the space $\{f^\star\}\times\mathcal{G}$. The resulting Stage-2 program $g^\star$ yields the Adaptive Pair Weighting (APW) objective: $\mathcal{L}_{APW}(x;f^\star,g^\star)=\frac{\sum_{i=1}^{M_{x}}g^\star(v_{i})f^\star(u_{i})}{\sum_{i=1}^{M_{x}}g^\star(v_{i})+\epsilon}$. This stage focuses exclusively on how much to learn from each pair within the set context.
\end{itemize}

By searching one program against a fixed counterpart, conditional search significantly reduces simultaneous design choices while ensuring the final APW objective benefits from the composed strengths of both programs. We acknowledge that this staged decoupling is a pragmatic compromise for computational efficiency. Because the search operates greedily—fixing $f^\star$ before observing its potential interactions with all possible weighting programs in $\mathcal{G}$—it may converge to a strong local optimum rather than the true global optimum of the joint space $\mathcal{F} \times \mathcal{G}$. Exploring more advanced co-evolutionary strategies that capture complex synergies between the two programs remains an important direction for future work.

\subsection{LLM-Driven Proposal and Behavioral Screening}
\label{sec:search-loop}

While the staged conditional search drastically reduces the dimensionality of the objective space, the LLM generator will inevitably propose candidates that are mathematically invalid or logically flawed for NCO tasks. Evaluating these arbitrary proposals blindly via NCO training would rapidly exhaust the computational budget. To circumvent this, we implement a rigorous behavioral gatekeeper. Before a candidate program is allocated any policy-training budget, it must successfully pass four groups of diagnostic checks:
\begin{enumerate}
    \setlength{\itemsep}{0pt}
    \setlength{\parskip}{0pt}

    \item \textbf{Executable Validity:} The compiled program is executed on a dummy batch for one training-like forward--backward pass under a predefined time limit. Because the loss is evaluated at every optimization step, an excessively expensive program would substantially reduce training throughput even if it were mathematically valid. This gate excludes candidates with unsupported operations, invalid outputs, detached losses, non-finite values or gradients, and unacceptably long runtimes before costly training begins.

    \item \textbf{Preference Consistency:} Verifies that the program respects the fundamental logic of comparison-based learning. Specifically, swapping the preferred and less-preferred solutions must reverse the signs of the loss gradients with respect to their log probabilities.

    \item \textbf{Scale Invariance:} This criterion checks that the program's behavior remains stable under a positive affine transformation of all solution costs by requiring
    $
    \left|\mathcal{L}(a o_i+b,\,a o_j+b)-\mathcal{L}(o_i,o_j)\right|
    \leq \epsilon_{\mathrm{aff}}, a>0.
    $
    Since this transformation preserves the solution ordering, the check prevents the objective from overfitting to the absolute cost magnitudes of a specific problem scale.

    \item \textbf{Weighting Non-degeneracy:} For each instance $x$, we compute $\operatorname{CV}_x(g)=     \frac{\operatorname{std}_{i=1,\ldots,M_x} g(v_i)}     {\operatorname{mean}_{i=1,\ldots,M_x}\vert{}g(v_i)\vert{}}.$We require $\operatorname{CV}_x(g)$ to exceed a predefined threshold, thereby excluding weighting programs that are effectively indistinguishable from the uniform weighting already evaluated in Stage 1.
\end{enumerate}
Programs failing any of these diagnostic gates are immediately excluded.
By treating objective design not just as a code generation task, but as a heavily constrained domain-specific search, this selective filtering serves as a powerful prior—reserving the expensive downstream training budget exclusively for programs that are not only executable, but also behaviorally admissible as preference objectives.

\subsection{Short-Horizon Evaluation and Diversity-Aware Selection}
\label{sec:fitness-selection}

Because preference objectives operate on solution sets sampled dynamically from the current policy, the distribution of pairwise comparisons is inherently non-stationary, shifting significantly as the policy evolves. Evaluating a candidate program solely from a random initialization heavily biases the search toward objectives that only accelerate early-stage learning, while executing a full-horizon training run for every survivor is computationally prohibitive.

To capture a robust performance profile while maintaining efficiency, we evaluate each behaviorally valid program using a fixed, shortened training budget across two distinct policy states: an \emph{early state} (randomly initialized) and a \emph{late state} (resumed from a fixed, partially trained checkpoint). Let $J_{c}^{s}$ denote the downstream validation cost achieved by candidate program $c$ when trained from state $s \in \{\text{early}, \text{late}\}$. To provide a stable relative metric, we establish a baseline cost $J_{\text{ref}}^{s}$ using the PO4COPs reference objective under identical conditions (state, data, and budget). The fitness of candidate $c$ is formally defined by its average cost improvement over this reference:
\[
F(c)=\frac{1}{2}\!\sum_{s\in
\{\mathrm{early},\mathrm{late}\}}
\left(J_{\mathrm{ref}}^s-J_c^s\right).
\]
By anchoring the fitness score against a matched baseline across distinct learning phases, this dual-state evaluation forces AutoPref to favor robust objectives that provide sustained optimization signals throughout the entire NCO training trajectory.

Finally, to translate these fitness scores into effective parent selection without inducing evolutionary stagnation, we actively balance downstream performance with structural diversity. Relying solely on raw fitness tends to cause premature convergence: once a high-scoring program pattern is found, subsequent LLM proposals often degenerate into minor syntactic variations, wasting the evaluation budget on redundant ideas. During survivor selection, we therefore filter out structurally redundant candidates, maintaining competing objective paradigms in the population. This allows AutoPref to continuously explore novel mechanisms while refining promising designs.
The effect of this structural-diversity filter is evaluated in Appendix~D.4.

\section{Experiments}
\label{sec:discovery-protocol}

\subsection{Experimental Setup}
\paragraph{Benchmark setups.}
We evaluate preference-objective discovery across four COP families: TSP, CVRP, FFSP, and JSSP.

Objective discovery is performed on TSP100, CVRP100, FFSP100, and JSSP15$\times$15. We refer to these problem scales as the \textbf{discovery scales}. After search, we freeze each discovered objective and use it unchanged for full-budget policy training at its discovery scale and at two other scales excluded from objective search, which we call \textbf{held-out scales}: TSP50/TSP1000, CVRP50/CVRP1000, FFSP50/FFSP1000, and JSSP10$\times$10/JSSP50$\times$20.

\paragraph{Baselines.}
We compare AutoPref with three groups of methods. (1) \textbf{Reference Solvers:} These provide the optimal or near-optimal costs used to compute the reported gaps. We use Concorde~\citep{applegate2006traveling}, HGS~\citep{vidal2022hgs}, and LKH3~\citep{helsgaun2017lkh3} for routing, alongside CP-SAT~\citep{ortools} for scheduling. (2) \textbf{Standard Neural Baselines:} This group includes AM~\citep{kool2019attention}, POMO~\citep{kwon2020pomo}, and SymNCO~\citep{kim2022symnco} for routing, MatNet~\citep{kwon2021matnet} for FFSP, and MGL~\citep{pmlr-v267-liao25a} for JSSP. (3) \textbf{Preference-based and Self-labeling Methods:} This category encompasses manually designed preference objectives—PO4COPs~\citep{pmlr-v267-pan25e} and BOPO~\citep{pmlr-v267-liao25a}—as well as SLIM~\citep{corsini2024self}, which iteratively trains on the best sampled solution. The two AutoPref variants differ in pair weighting: UPW assigns uniform weights to the discovered pairwise loss, whereas APW weights the same loss using the discovered set-aware program. For a fair comparison, all methods use the same task-specific neural architecture and training budget within each problem family. Appendix~B provides scale-specific configurations.

\paragraph{Metrics.}

We evaluate model performance using three primary metrics: mean solution cost, optimality gap, and total evaluation time. Given the mean solution cost $C$ and the problem-specific reference cost $C_{ref}$ over a shared test set, the optimality gap is defined as $\mathrm{Gap}=\frac{C-C_{\mathrm{ref}}}{C_{\mathrm{ref}}}\times 100\%$. For both cost and gap, lower values indicate better performance. Finally, time is the total runtime over the full test set.

\subsection{Main Results}
\label{sec:evaluation}

While our short-horizon evaluation during search efficiently identifies promising objectives, a critical question remains: do these short-horizon advantages persist under full-budget training and generalize to unseen scales? To answer this, we evaluate the frozen discovered objectives under full-budget policy training across both discovery and held-out scales.

\paragraph{Routing.}

\begin{table*}[t]
\caption{Routing results at the discovery scale (D) and two held-out scales (H). Test sets contain 10,000 instances for TSP50, TSP100, CVRP50, and CVRP100; 100 instances for TSP1000; and 128 instances for CVRP1000. Time is the total runtime over the full test set. The TSP1000 reference uses one exact Concorde solve per instance, and the CVRP1000 reference reports the best HGS result over three seeds. Bold marks the best result among PO4COPs, BOPO, SLIM, UPW, and APW.}
\label{tab:routing-results}
\centering
\resizebox{\textwidth}{!}{
\begin{tabular}{l*{18}{c}}
\toprule
Method
& \multicolumn{3}{c}{TSP100 (D)}
& \multicolumn{3}{c}{TSP50 (H)}
& \multicolumn{3}{c}{TSP1000 (H)}
& \multicolumn{3}{c}{CVRP100 (D)}
& \multicolumn{3}{c}{CVRP50 (H)}
& \multicolumn{3}{c}{CVRP1000 (H)} \\
\cmidrule(lr){2-4}\cmidrule(lr){5-7}\cmidrule(lr){8-10}
\cmidrule(lr){11-13}\cmidrule(lr){14-16}\cmidrule(lr){17-19}
& Cost $\downarrow$ & Gap (\%) $\downarrow$ & Time (s)
& Cost $\downarrow$ & Gap (\%) $\downarrow$ & Time (s)
& Cost $\downarrow$ & Gap (\%) $\downarrow$ & Time (s)
& Cost $\downarrow$ & Gap (\%) $\downarrow$ & Time (s)
& Cost $\downarrow$ & Gap (\%) $\downarrow$ & Time (s)
& Cost $\downarrow$ & Gap (\%) $\downarrow$ & Time (s) \\
\midrule
\multicolumn{19}{l}{\textit{Reference solvers}}\\
Concorde/HGS
& 7.7609 & 0.000 & $2.4\times 10^3$
& 5.6903 & 0.000 & 798
& 23.1143 & 0.000 & $1.62{\times}10^{4}$
& 15.5857 & 0.000 & $1.29{\times}10^{4}$
& 10.3659 & 0.000 & $7.5\times 10^3$
& 119.1273 & 0.000 & $2.30{\times}10^{5}$ \\
LKH3
& 7.7609 & 0.000 & $1.1\times 10^3$
& 5.6903 & 0.000 & 120
& 23.1143 & 0.000 & $7.8\times 10^3$
& 15.6673 & 0.524 & $7.2\times 10^3$
& 10.3670 & 0.011 & $3.6\times 10^3$
& 123.2142 & 3.431 & $8.2\times 10^3$ \\
\midrule
\multicolumn{19}{l}{\textit{Standard neural solvers}}\\
AM
& 8.1233 & 4.669 & 5
& 5.7940 & 1.823 & 2
& 30.4472 & 31.724 & 378
& 16.5804 & 6.382 & 5
& 10.9490 & 5.625 & 2
& 131.0755 & 10.030 & 266 \\
POMO
& 7.7752 & 0.185 & 64
& 5.6968 & 0.114 & 15
& 31.2066 & 35.001 & 370
& 15.7604 & 1.121 & 77
& 10.4630 & 0.937 & 47
& 127.0254 & 6.623 & 224 \\
SymNCO
& 7.7746 & 0.177 & 63
& 5.6976 & 0.129 & 15
& 30.1819 & 30.600 & 366
& 15.8180 & 1.491 & 78
& 10.4617 & 0.924 & 48
& 128.1365 & 7.563 & 251 \\
\midrule
\multicolumn{19}{l}{\textit{Preference-objective and self-labeling baselines}}\\
PO4COPs
& 7.7744 & 0.174 & 64
& 5.6982 & 0.139 & 15
& 25.5745 & 10.644 & 491
& 15.8116 & 1.449 & 77
& 10.4904 & 1.201 & 47
& 127.4277 & 6.968 & 237 \\
SLIM
& 7.7733 & 0.160 & 64
& 5.7005 & 0.179 & 15
& 28.7467 & 24.368 & 498
& 15.7779 & 1.233 & 77
& 10.4696 & 1.000 & 47
& 136.0070 & 14.169 & 261 \\
BOPO
& 7.8500 & 1.148 & 64
& 5.7028 & 0.220 & 15
& 27.2378 & 17.840 & 496
& 16.0818 & 3.183 & 77
& 10.5438 & 1.716 & 47
& 127.6575 & 7.161 & 207 \\
\midrule
\multicolumn{19}{l}{\textit{Discovered preference objectives}}\\
UPW
& 7.7706 & 0.125 & 64
& 5.6920 & 0.030 & 15
& 25.3580 & 9.708 & 497
& 15.7550 & 1.086 & 77
& 10.4503 & 0.814 & 47
& 123.6104 & 3.763 & 250 \\
APW
& \textbf{7.7677} & \textbf{0.088} & 64
& \textbf{5.6914} & \textbf{0.019} & 15
& \textbf{25.2858} & \textbf{9.395} & 490
& \textbf{15.7450} & \textbf{1.022} & 77
& \textbf{10.4370} & \textbf{0.686} & 47
& \textbf{123.1858} & \textbf{3.407} & 198 \\
\bottomrule
\end{tabular}}
\end{table*}

On routing tasks (Table~\ref{tab:routing-results}), APW consistently establishes a new state-of-the-art among preference-based methods at both the discovery and held-out scales. Most notably, APW’s performance margin over baselines expands significantly on larger, held-out scale instances. This trend highlights a critical vulnerability in hand-designed objectives: signal dilution. BOPO averages trajectory log-likelihood over the number of construction steps, inevitably weakening the contribution of each step as the time horizon grows. Conversely, PO4COPs applies uniform weighting across all pairs, allowing weak or near-tie comparisons to drown out informative ones. APW effectively counteracts both failure modes by preserving accumulated trajectory-level signals and adaptively assigning larger weights to the most informative comparisons. We further analyze the specific form of the discovered TSP APW objective and the interaction between its two programs in Section~\ref{sec:analysis}.

\paragraph{Scheduling.}

\begin{table*}[!t]
\caption{Scheduling results at the discovery scale (D) and two held-out scales (H). Test sets contain 1,000 instances for FFSP50 and FFSP100, and 100 instances for FFSP1000, JSSP10$\times$10, JSSP15$\times$15, and JSSP50$\times$20. Time is the total runtime over the full test set. Bold marks the best result among PO4COPs, BOPO, SLIM, UPW, and APW.}
\label{tab:scheduling-results}
\centering
\resizebox{\textwidth}{!}{
\begin{tabular}{l*{18}{c}}
\toprule
Method & \multicolumn{3}{c}{FFSP100 (D)}
& \multicolumn{3}{c}{FFSP50 (H)}
& \multicolumn{3}{c}{FFSP1000 (H)}
& \multicolumn{3}{c}{JSSP15$\times$15 (D)}
& \multicolumn{3}{c}{JSSP10$\times$10 (H)}
& \multicolumn{3}{c}{JSSP50$\times$20 (H)} \\
\cmidrule(lr){2-4}\cmidrule(lr){5-7}\cmidrule(lr){8-10}
\cmidrule(lr){11-13}\cmidrule(lr){14-16}\cmidrule(lr){17-19}
& Cost $\downarrow$ & Gap (\%) $\downarrow$ & Time (s)
& Cost $\downarrow$ & Gap (\%) $\downarrow$ & Time (s)
& Cost $\downarrow$ & Gap (\%) $\downarrow$ & Time (s)
& Cost $\downarrow$ & Gap (\%) $\downarrow$ & Time (s)
& Cost $\downarrow$ & Gap (\%) $\downarrow$ & Time (s)
& Cost $\downarrow$ & Gap (\%) $\downarrow$ & Time (s) \\
\midrule
\multicolumn{19}{l}{\textit{Reference solvers}}\\
CP-SAT
& 90.941 & 0.000 & $1.43\times 10^5$
& 49.440 & 0.000 & $5.86 \times 10^4$
& 803.590 & 0.000 & $5.50{\times}10^{5}$
& 1205.44 & 0.000 & $3.54{\times}10^{4}$
& 822.26 & 0.000 & 219
& 2882.22 & 0.000 & $8.00{\times}10^{5}$ \\
\midrule
\multicolumn{19}{l}{\textit{Standard neural solvers}}\\
MatNet/MGL
& 89.960 & -1.079 & 1113
& 49.845 & 0.819 & 348
& 804.380 & 0.098 & 1105
& 1352.57 & 12.206 & 65
& 892.01 & 8.483 & 36
& 3210.29 & 11.383 & 305 \\
\midrule
\multicolumn{19}{l}{\textit{Preference-objective and self-labeling baselines}}\\
PO4COPs
& 89.680 & -1.387 & 1113
& 49.769 & 0.665 & 348
& 804.380 & 0.098 & 974
& 1649.54 & 36.841 & 65
& 1070.09 & 30.140 & 36
& 3112.68 & 7.996 & 427 \\
SLIM
& 89.570 & -1.508 & 1204
& 49.716 & 0.558 & 398
& 804.460 & 0.108 & 983
& 1874.54 & 55.507 & 65
& 1092.54 & 32.870 & 36
& 3112.68 & 7.996 & 440 \\
BOPO
& 89.672 & -1.395 & 1113
& 49.755 & 0.637 & 413
& 805.670 & 0.259 & 1098
& 1303.18 & 8.108 & 65
& 855.78 & 4.077 & 36
& 3104.15 & 7.700 & 429 \\
\midrule
\multicolumn{19}{l}{\textit{Discovered preference objectives}}\\
UPW
& 89.420 & -1.673 & 1113
& 49.730 & 0.587 & 348
& 804.380 & 0.098 & 1126
& 1305.49 & 8.300 & 65
& 851.21 & 3.521 & 36
& 3103.29 & 7.670 & 423 \\
APW
& \textbf{89.380} & \textbf{-1.716} & 1113
& \textbf{49.700} & \textbf{0.526} & 348
& \textbf{804.160} & \textbf{0.071} & 919
& \textbf{1297.40} & \textbf{7.629} & 65
& \textbf{847.32} & \textbf{3.048} & 36
& \textbf{3102.44} & \textbf{7.641} & 446 \\
\bottomrule
\end{tabular}}
\end{table*}

For scheduling problems (Table~\ref{tab:scheduling-results}), APW similarly achieves the lowest solution costs. Unlike routing tasks, scheduling problems such as FFSP and JSSP involve inherently complex, multi-dimensional constraints, making objective optimization particularly challenging. Despite this, APW consistently outperforms both standard neural solvers and manually designed preference baselines. Most notably, whether evaluated on the initial discovery scale or tested on larger held-out scales, the discovered APW objective robustly maintains its performance advantage, underscoring AutoPref's exceptional resilience and broad applicability across diverse NCO domains.

\paragraph{The Necessity of Task-Specific Discovery.}

Viewed holistically, the empirical results decisively validate the need for automated objective discovery. Hand-designed preference objectives frequently struggle or degrade performance on specific tasks; for instance, BOPO underperforms across several routing settings, while PO4COPs significantly degrades performance on JSSP. In stark contrast, APW consistently achieves the lowest costs across all twelve evaluated settings. Under identical neural architectures and training budgets, APW delivers the strongest results among all preference-based NCO objectives, proving that learning objectives should be treated as task-adaptive components rather than static, hand-crafted ones.

\subsection{How Do the Two Programs Contribute?}

Having established AutoPref's consistent performance gains across problem families, we next isolate the specific roles of the pairwise loss and set-aware weighting programs. To achieve this, we conduct a factorial ablation, evaluating four configurations formed by crossing the baseline pairwise loss (PO4COPs) and the discovered pairwise loss with either uniform or discovered weighting schemes. Table~\ref{tab:factorial-ablation} reports the absolute cost reduction relative to the fully manual baseline (PO4COPs loss with uniform weighting), where positive values indicate performance improvements.

\begin{table}[t]
\caption{Factorial ablation of the two objective programs.}
\label{tab:factorial-ablation}
\centering
\scriptsize
\begin{tabular}{llcc}
\toprule
Pairwise loss & Weighting & TSP100 & FFSP100 \\
\midrule
PO4COPs & Uniform & 0.0000 & 0.000 \\
Discovered & Uniform & 0.0039 & 0.260 \\
PO4COPs & Discovered & 0.0011 & 0.201 \\
Discovered & Discovered & \textbf{0.0067} & \textbf{0.300} \\
\bottomrule
\end{tabular}
\end{table}

The ablation clearly demonstrates the effectiveness of both discovered programs across different problem families. On both TSP100 and FFSP100, replacing the manual baseline with either the discovered pairwise loss or the discovered set-aware weighting program individually yields consistent performance improvements. Crucially, combining the two discovered components achieves the highest overall cost reduction on both tasks. This confirms that the two programs capture beneficial optimization signals, and that their joint application successfully maximizes the performance of preference-based NCO. We further analyze whether these gains persist throughout training and whether staged conditional search is more effective than joint search in Appendices~D.1 and~D.2, respectively.

\subsection{What Does a Discovered Objective Learn?}
\label{sec:analysis}

To demystify how the discovered pairwise loss and weighting programs shape policy learning, we mathematically analyze the APW objective discovered for TSP. 

Appendix~E reports the objectives discovered for all problem families.

\textbf{The Discovered Objective.} The objective combines a bounded, cost-calibrated logistic loss with a set-aware normalization weight. 

For a comparison $i$, let $\Delta p_i=p_w^{(i)}-p_l^{(i)}$ denote the log-probability margin and $\Delta o_i=o_l^{(i)}-o_w^{(i)}$ the cost gap. For sampled solution set $S_x$, let $m_o(S_x)$ denote its cost median absolute deviation and $\sigma_p(S_x)$ its log-probability standard deviation. 
Utilizing fixed constants $\alpha_{\mathrm T}>0$, $0<\beta_{\mathrm T}<1$, and $C>0$, the objective is:
{\scriptsize
\begin{align*}
q_i&=\frac{\Delta o_i}{1+|\Delta o_i|}, \,\,
s_{\mathrm T}^{(i)}
=-\log\sigma\!\left(\operatorname{clip}_{[-C,C]}\!\left[
\alpha_{\mathrm T}\Delta p_i
\left(1-\beta_{\mathrm T} \cdot q_i \right)\right]\right)\\
\omega_{\mathrm T}^{(i)}
&=\operatorname{clip}_{[\ell,u]}\left[\frac{\Delta o_i}{m_o(S_x)+\varepsilon}
\cdot\frac{|\Delta p_i|}{\sigma_p(S_x)+\varepsilon}\right], \,\,
\mathcal L_{\mathrm T}
=\frac{\sum_i\omega_{\mathrm T}^{(i)}s_{\mathrm T}^{(i)}}
{\sum_i\omega_{\mathrm T}^{(i)}+\varepsilon}.
\label{eq:tsp-case-objective}
\end{align*}
}

\textbf{Mechanism Analysis.} The mathematical structure of this objective reveals a departure from human expert priors. Within \(s_i\), the factor \(1-\beta_{\mathrm T}q_i\) acts as a cost-gap-dependent inverse temperature. Because \(q_i=\Delta o_i/(1+|\Delta o_i|)\) strictly increases with the cost gap, a larger cost gap reduces this inverse temperature and moves \(\alpha_{\mathrm T}(1-\beta_{\mathrm T}q_i)\Delta p_i\) toward zero. When \(\Delta p_i>0\), this prevents premature sigmoid saturation; when \(\Delta p_i<0\), it robustly bounds the loss from exploding. Notably, this behavior directly contradicts the expert prior encoded in BOPO, which aggressively magnifies \(\Delta p_i\) by multiplying it by \(o_l^{(i)}/o_w^{(i)}=1+\Delta o_i/o_w^{(i)}\). Simultaneously, the discovered set-aware weighting program counterbalances this effect. Because \(\omega_i\) depends on both \(\Delta o_i/m_o(S_x)\) and \(|\Delta p_i|/\sigma_p(S_x)\), a larger cost gap assigns a significantly higher global weight to the comparison even as its local margin is tempered by the inverse temperature. This synergy allows large-gap comparisons to exert a stronger overall influence on the policy update without subjecting the model to extreme, destabilizing individual gradient steps.

\textbf{Behavioral Analysis.} 

\begin{figure}[t]
\centering
\begin{minipage}[t]{0.485\columnwidth}
\centering
\includegraphics[width=\linewidth]{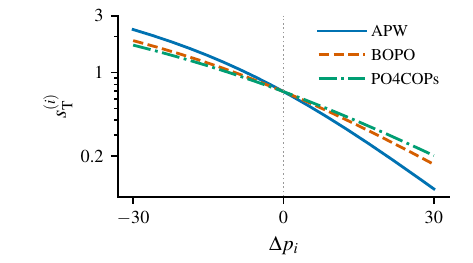}
{\scriptsize (a) Loss vs. margin.\par}
\end{minipage}
\hfill
\begin{minipage}[t]{0.485\columnwidth}
\centering
\includegraphics[width=\linewidth]{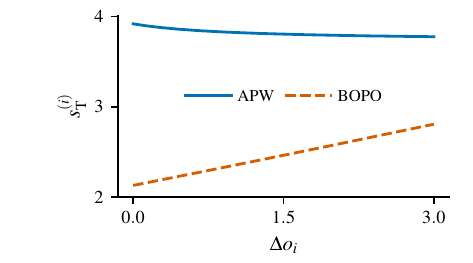}
{\scriptsize (b) Loss vs. cost gap.\par}
\end{minipage}

\begin{minipage}[t]{0.485\columnwidth}
\centering
\includegraphics[width=\linewidth]{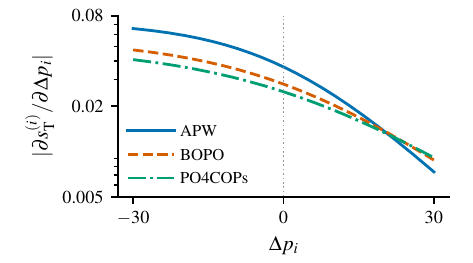}
{\scriptsize (c) Gradient vs. margin.\par}
\end{minipage}
\hfill
\begin{minipage}[t]{0.485\columnwidth}
\centering
\includegraphics[width=\linewidth]{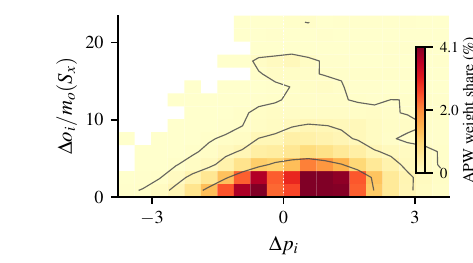}
{\scriptsize (d) APW weight distribution.\par}
\end{minipage}

\caption{Optimization behavior of the discovered TSP preference objective on pairwise comparisons sampled from a TSP100 training batch.}
\label{fig:tsp-apw-loss-analysis}
\end{figure}

Figure~\ref{fig:tsp-apw-loss-analysis} corroborates these mechanics empirically on sampled TSP100 training samples.
Panel~(a) shows that APW changes the loss most sharply as \(\Delta p_i\) moves from negative to positive. Panel~(b) reveals the opposite cost-gap responses: for a fixed negative margin, the APW loss decreases with \(\Delta o_i\), whereas the BOPO loss increases. Panel~(c) shows that APW produces the strongest gradients for negative and near-zero margins, emphasizing incorrectly ordered and ambiguous comparisons. Panel~(d) shows that most APW weight lies in the frequently sampled region with small-to-moderate normalized cost gaps and near-zero to small positive margins. Together, these results show that APW strongly corrects misordered comparisons while concentrating its total weight on the comparisons most often encountered during training.

\section{Related Work}
\label{sec:related-work}
\paragraph{Conventional constructive NCO paradigms.}

Constructive neural solvers generate multiple solutions per instance during training or instance-specific search. Conventional paradigms leverage these samples through shared policy-gradient baselines, as in POMO~\citep{kwon2020pomo}; symmetry exploitation, as in SymNCO~\citep{kim2022symnco}; instance-specific adaptation, as in active search~\citep{hottung2022efficient}; or pseudo-labeling for self-improvement~\citep{pirnay2024self}. However, these approaches largely overlook the rich relational structure encoded by the cost ordering within each sampled set.

\paragraph{Preference optimization for NCO.}

Bridging this structural gap, preference-based NCO utilizes relative solution quality for direct policy supervision. Foundational works like PO4COPs~\citep{pmlr-v267-pan25e} employ fixed pairwise objectives with uniform weighting, while BOPO~\citep{pmlr-v267-liao25a} relies on best-anchored pairs scaled by hand-crafted cost ratios. Recent extensions like POCCO~\citep{fan2025pocco} and UCPO~\citep{fang2026ucpo} adapt these concepts for multi-objective and constrained scenarios. Despite validating the efficacy of preference-based training, these methods are intrinsically bottlenecked by manual engineering.

\paragraph{Automated objective discovery.}
The paradigm of automated discovery seeks to replace manual engineering by treating algorithmic components as search targets. Prior work learns optimization schedules~\citep{xu2019autoloss} or parameterized losses~\citep{bechtle2021meta}, searches explicit loss~\citep{Li_2022_CVPR} or optimizer expressions~\citep{chen2023symbolic}, and uses LLMs to propose executable mathematical programs~\citep{romera2024funsearch}, combinatorial heuristics~\citep{pmlr-v235-liu24bs}, or reward functions~\citep{ma2024eureka}. AutoPref is the first to bring LLM-guided automated discovery to preference-objective design in preference-based NCO. 

\section{Conclusion}

We formulate the design of preference objectives for NCO as an automated discovery problem over a coupled programmatic space. By factorizing the objective into a pairwise loss program and a set-aware weighting program, AutoPref transforms static, manual design choices into explicit, searchable components that naturally subsume existing manually designed preference objectives. To overcome the computational bottleneck of objective search, we introduce an efficient staged conditional search strategy with behavioral gates. Empirically, the discovered objectives establish a new state of the art for preference-based NCO across four distinct COP families, generalizing robustly to held-out problem scales under strictly matched protocols. These findings strongly advocate for a paradigm shift: learning objectives should be treated as task-adaptive components to be computationally discovered, rather than static rules repeatedly specified by experts. Future work will focus on extending this paradigm by exploring loss formulations beyond pairwise objectives and leveraging surrogate models to accelerate objective evaluation.

\bibliography{references}
\end{document}